\documentclass{article}

\usepackage[final]{corl_2018}

\usepackage{algorithm}
\usepackage{algpseudocode}
\usepackage{subfigure}
\usepackage{caption}
\usepackage{multicol}
\usepackage[pdftex]{graphicx}
\usepackage{array,multirow}
\usepackage{tabularx}
\usepackage{adjustbox}
\usepackage{enumerate}

\graphicspath{{../pdf/}{../jpeg/}{./image/}}    
\DeclareGraphicsExtensions{.pdf,.jpeg,.png,.jpg}  

\newcommand{\eref}[1]{(\ref{#1})}

\usepackage{amsmath}
\usepackage{amssymb}

\usepackage{setspace}
\let\Algorithm\algorithm
\renewcommand\algorithm[1][]{\Algorithm[#1]\setstretch{1.4}}
\usepackage{enumitem,kantlipsum}

\usepackage{titlesec}
 \titlespacing*{\section}
 {0pt}{0.5ex plus 1ex minus .2ex}{0.3ex plus .2ex}
 \titlespacing*{\subsection}
 {0pt}{0.2ex plus 1ex minus .2ex}{0.1ex plus .2ex}
\usepackage{caption}
\title{Including Uncertainty when \\ Learning from Human Corrections}

\author{
  Dylan P. Losey\\
  Rice University\\
  \texttt{dlosey@rice.edu} \\
  \And
  Marcia K. O'Malley\\
  Rice University\\
  \texttt{omalleym@rice.edu}
}

\begin{document}
\maketitle


\begin{abstract} It is difficult for humans to efficiently teach robots how to correctly perform a task. One intuitive solution is for the robot to iteratively learn the human's preferences from corrections, where the human improves the robot's current behavior at each iteration. When learning from corrections, we argue that while the robot should estimate the most likely human preferences, it should also \emph{know what it does not know}, and integrate this uncertainty as it makes decisions. We advance the state-of-the-art by introducing a Kalman filter for learning from corrections: this approach obtains the uncertainty of the estimated human preferences. Next, we demonstrate how the estimate uncertainty can be leveraged for \emph{active learning} and \emph{risk-sensitive} deployment. Our results indicate that obtaining and leveraging uncertainty leads to faster learning from human corrections.
\end{abstract}



\section{Introduction} \label{sec:intro}

While robots can be pre-programmed by an expert designer to execute a wide range of behaviors, each robot user has different preferences for how their robot should behave. Recent work has focused on learning the human end-user's preferences from \emph{corrections}: here the robot shows the human how it has been pre-programmed to perform the task, and the human corrects the robot's behavior to suit their personal preferences. Importantly, these corrections do not need to be perfect; instead, a human correction is simply a \emph{noisy improvement} of the robot's current behavior.

\begin{figure}[t]

	\begin{center}
		\includegraphics[width=1\columnwidth]{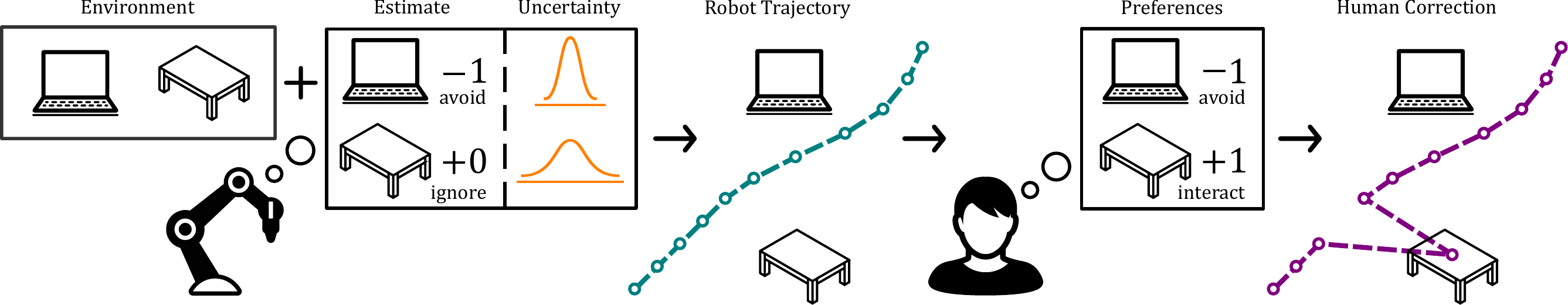}
        
        \vspace{0.5em}

		\caption{Iterative learning from human corrections. Given an environment and the current estimated preferences, the robot selects a trajectory (teal). The human then observes this trajectory, and provides a correction to better match their true preferences (purple). Traditionally, the robot uses the correction to update its estimate at the next iteration. We propose that the robot should also obtain the uncertainty over this estimate (orange).}

		\label{fig:intro}
	\end{center}

\end{figure}

Consider a robotic manipulator carrying a cup of coffee for its human end-user. This robot knows to avoid obstacles, but is not sure about the human's preferences: e.g., should the robot carry coffee over a laptop, across a table, or avoid both regions? When learning from corrections, the robot shows the human its current estimate of the optimal trajectory. The human then corrects this trajectory---using physical human-robot interaction or a virtual interface---and, for example, pushes the robot farther away from the laptop. The robot learns iteratively (i.e., online), and updates its understanding of the human's preferences after each correction. See Fig.~\ref{fig:intro} for an overview of this process.

Human corrections indicate which preferences are more probable. For instance, if the human pushes the robot away from the laptop, then the robot should infer that preferences which result in the robot avoiding the laptop are more likely. Within the state-of-the-art, robots estimate the \emph{most likely} human preferences given the human's corrections \cite{ratliff2006, shivaswamy2015}. However, these robots miss out on the \emph{uncertainty} of their estimate: i.e., in practice, the robot may not understand the human's preferences with much confidence. Our insight is that---because human corrections imply a probability distribution over their preferences---\emph{the robot should not only estimate the most likely human preferences from these corrections, but should also recognize which estimates it is not confident about}.

Let us return to our working example, where a user wants the robot to avoid carrying coffee over their laptop. Because the human pushed the robot away from their laptop---and the table is nearby---the robot learns to avoid both the laptop and table. If the robot only estimates the most likely human preferences, it will avoid carrying coffee over the table. But the human may actually want the robot to move over the table! A robot which knows the uncertainty of its estimate is \emph{confident} that it should avoid the laptop, but \emph{unsure} whether it should avoid the table. We can leverage this uncertainty to elicit informative corrections (that teach the robot about the human's table preference) or for risk-sensitive deployment (that avoids the table entirely) when more corrections are unavailable.

\textbf{Contributions}. First, we show how iterative inverse reinforcement learning can be performed using a \emph{Kalman filter}, where human corrections are noisy observations. This approach extends the state-of-the-art to now track the uncertainty over the estimated preferences. Next, we leverage uncertainty within our setting to \emph{actively learn} from human corrections, so that the robot can elicit corrections that will reduce uncertainty. After the learning is completed, and the human stops providing corrections, we also describe how uncertainty can be leveraged for \emph{risk-sensitive deployment}, i.e., to avoid interacting with preferences about which the robot is most uncertain.

\section{Related Work}

The problem we are considering is based on learning from human corrections, an instance of inverse reinforcement learning (IRL). Our solution will also build upon active learning approaches, where the robot reasons over uncertainty during IRL. Here we briefly overview both fields.

\noindent \textbf{Inverse reinforcement learning}. Also known as inverse optimal control, IRL attempts to recover the human's preferences from demonstrations which are optimal \cite{abbeel2004,ng2000,osa2018}. In practice, however, it is challenging for humans to provide optimal demonstrations: consider an end-user trying to guide the motion of a multi degree-of-freedom (DoF) robotic manipulator \cite{akgun2012}. One solution is presented by probabilistic IRL approaches \cite{ziebart2008,ramachandran2007}, which assume that the human is noisily optimal, and learn a distribution over the space of possible human preferences. 

Alternatively, the robot can learn from corrections. At each iteration the robot maximizes its current estimate of the human's preferences, and the human responds by slightly improving, or correcting, the robot's behavior. Here the human's corrections do not need to be noisily optimal. Shivaswamy and Joachims \cite{shivaswamy2015} model learning from corrections as Coactive Learning, and derive iterative IRL algorithms which are similar to \cite{ratliff2006}. In particular, the Preference Perceptron from \cite{shivaswamy2015} has been applied to robotic manipulators by \cite{bajcsy2017,bajcsy2018,jain2015}. While these works learn a maximum a posteriori estimate of the human's objective, we note that they do not maintain the uncertainty over this estimate.

\noindent \textbf{Active learning}. Other works explicitly reason over uncertainty while learning from the human. For instance, active learning reduces uncertainty by enabling the robot to choose informative queries, which are then answered by the expert human \cite{settles2012}. Active learning has previously been applied to improve IRL in \cite{lopes2009, cohn2011, cui2018}. Most relevant to our work is recent research by Cui and Niekum \cite{cui2018}, where the robot proposes a trajectory to the human, and the human segments this trajectory into ``good'' and ``bad'' portions. The robot updates its understanding of the human's preferences based on this segmentation; moreover, to increase its learning rate, the robot actively generates trajectories which are expected to result in user critiques that best reduce uncertainty. Our research is similar to \cite{cui2018}, but here we focus on learning from human corrections rather than user segmentation.

\section{Background}

Within this section we derive the Preference Perceptron, the current state-of-the-art approach when learning from human corrections \cite{ratliff2006,shivaswamy2015,bajcsy2017}. We note that the Preference Perceptron is equivalent to online Maximum Margin Planning without any loss function.

\textbf{Notation}. Consider a robot with state $x \in \mathcal{X}$, action $a \in \mathcal{A}$, and dynamics $f$. These dynamics define the probability distribution over the robot's next state given its current state and action: i.e., $f(x^{i+1} | x^i, a^i)$, where $i$ denotes the current timestep. The task ends after $T \in \mathbb{Z}^+$ timesteps.

\textbf{Trajectory and Environment}. Let us define the robot's trajectory $\xi \in \Xi$ as the sequence of robot states $x$, such that $\xi = x^{0:T}$. Although this trajectory describes the robot's behavior, it does not tell us about the world in which the robot is acting. Accordingly, let $E$ denote the robot's environment; we can equivalently think of $E$ as the ``world description'' \cite{huang2017}, or as the ``context'' \cite{shivaswamy2015}. We include the prior distribution over the robot's start state $x^0$ within the environment $E$.

\textbf{Reward}. The human end-user has in mind a reward function, $R$, which determines the utility of the robot following trajectory $\xi$ in environment $E$. Like previous IRL works \cite{ratliff2006,abbeel2004,osa2018,ziebart2008}, we will assume that $R$ is a linear combination of features $\phi(\xi,E) \in [0,1]^k$ weighted by a parameter vector $\theta \in \mathbb{R}^k$:
\begin{equation} \label{eq:B1}
	R(\xi,E) = \theta \cdot \sum_{i=0}^T \phi(x^i,E) = \theta \cdot \phi(\xi,E)
\end{equation}
The features $\phi$ are known by both the human and the robot. Given $\theta$, we have described an instance of a Markov decision process \cite{puterman2014} that can be solved to find the optimal robot policy. In practice, however, the true reward parameter $\theta$ is \emph{known only by the human}. In other words, the choice of $\theta$ is user-specific: $\theta$ encodes the human's \emph{preferences} over the robot's trajectory, and varies from one end-user to another \cite{jain2015}. Hence, the robot must learn $\theta$ from the current end-user.

\textbf{Corrections}. The robot learns about the human's true preferences $\theta$ from the human's corrections. At each iteration $t$, the robot observes an environment $E^t$ and chooses a trajectory $\xi^t$. The human end-user observes both $E^t$ and $\xi^t$, and corrects the robot's trajectory to $\xi^t_h$. We assume that the human's corrected trajectory has higher reward than the robot's original trajectory:
\begin{gather} \label{eq:B2}
	R(\xi_h^t,E^t) > R(\xi^t,E^t) \\
\theta \cdot \phi(\xi^t_h,E^t) > \theta \cdot \phi(\xi^t,E^t)
\end{gather}
where we have applied the reward function \eref{eq:B1}. Intuitively, here we are claiming that the human sees the robot's behavior, and then modifies that behavior so that the robot's actions better align with their preferences. Notice that the human does not have to correct the robot to the optimal trajectory---i.e., provide a noisily optimal demonstration---but only needs to slightly \emph{improve} the robot's trajectory.

\textbf{Preference Perceptron}. Given that the human's correction returns an improved trajectory, the current state-of-the-art robot learns with the Preference Perceptron. Let $\hat{\theta}$ be the robot's estimate of the human's true preferences $\theta$. At each iteration $t$, the robot updates $\hat{\theta}$ to maximize the margin between the estimated rewards associated with $\xi^t$ and $\xi_h^t$ such that $\hat{\theta} \cdot \phi(\xi^t,E^t) < \hat{\theta} \cdot \phi(\xi^t_h,E^t)$. Put another way, the robot minimizes the following cost function:
\begin{equation} \label{eq:B4}
	J\big(\hat{\theta}\,\big) = \hat{\theta} \cdot \big[\phi(\xi^t,E^t) - \phi(\xi^t_h,E^t)\big]
\end{equation}
Since $J$ is differentiable with respect to $\hat{\theta}$, we leverage online gradient descent \cite{bottou1998} to get:
\begin{equation} \label{eq:B5}
	\hat{\theta}^{t+1} = \hat{\theta}^t + \alpha^t \big[\phi(\xi^t_h,E^t) - \phi(\xi^t,E^t)\big]
\end{equation}
where $\alpha > 0$ is the \emph{learning rate}. This update rule \eref{eq:B5} is referred to as the Preference Perceptron. Intuitively, a robot leveraging \eref{eq:B5} learns by comparing feature counts between the corrected and original trajectories: features that the human has increased are weighted more highly, while features that the human has decreased are weighted less highly. Prior work has demonstrated that \eref{eq:B5} is also the robot's \emph{maximum a posteriori} (MAP) estimate of the human's true preferences $\theta$ \cite{bajcsy2017}.

\textbf{Optimal Trajectory}. Given $\hat{\theta}^t$ and $E^t$, the robot can identify an optimal trajectory that maximizes its current estimate of the human's reward. We obtain this trajectory by solving:
\begin{equation}\label{eq:B6}
	\xi^t = \text{arg}\max_{\xi \in \Xi}  ~ \hat{\theta}^t \cdot \phi(\xi, E^t)
\end{equation}
For robotic manipulators, a trajectory optimizer such as \cite{karaman2011,schulman2014} can be leveraged to solve \eref{eq:B6}. 

\textbf{Summary}. The robot observes an environment $E^t$ at each iteration $t$. Based on $E^t$ and the robot's current estimate of $\theta$, the robot solves \eref{eq:B6} for its trajectory $\xi^t$. The human then corrects the robot's trajectory, and provides an improved trajectory $\xi_h^t$. Finally, the robot updates its estimate of $\theta$ using \eref{eq:B5}, and the process repeats at the next iteration. We can alternatively think of $\xi^t$ as the label which the robot assigns to the input $E^t$, while $\xi_h^t$ is an improved label provided by the human's correction.

\textbf{Uncertainty}. Although \eref{eq:B5} provides a maximum a posteriori estimate of $\theta$, this Preference Perceptron does not obtain a probability distribution over $\theta$. Thus, when the robot learns using (\ref{eq:B5}), we do not know the uncertainty of our estimate $\hat{\theta}$. Instead, a robot using the Preference Perceptron falsely assumes that it completely understands the human's preferences after each correction.

\section{Kalman Filter for Inverse Reinforcement Learning} \label{sec:kalman}

Our first contribution is to recognize that \eref{eq:B5}---the standard IRL update rule for learning from human corrections---can be rewritten as a Kalman filter \cite{choset2005}. We argue that the key advantage to using a Kalman filter is that it not only provides an iterative estimate similar to \eref{eq:B5}, but it also \emph{obtains the uncertainty over this estimate}. Here we explain how to apply a Kalman filter for iterative IRL.

\textbf{Transition Model}. We model the human's preferences $\theta^t$ as constant between iterations, like previous IRL works \cite{ratliff2006,abbeel2004,osa2018,ziebart2008}. Then, we can write the transition function:
\begin{equation} \label{eq:K1}
	\theta^{t} = \theta^{t-1} + m^t \quad \quad m^t \sim \mathcal{N}(0,M^t)
\end{equation}
where $m^t$ is the \emph{process noise} at iteration $t$. We assume that $m^t$ is drawn from a zero-mean Gaussian distribution with covariance $M^t$. Introducing this process noise enables the robot to capture how the imperfect human may unintentionally alter their preferences between iterations, i.e., the human may not know exactly what they want: \eref{eq:K1} implies that the end-user's preferences are \emph{noisily constant}.

\textbf{Observation Model}. The robot learns about the end-user's preferences $\theta^t$ by observing the human's corrected trajectory $\xi^t_h$, and---more specifically---by observing the features $\phi$ along that corrected trajectory. Accordingly, the robot has an observation model:
\begin{equation} \label{eq:K2}
	\phi(\xi_h^t,E^t) = \phi\big(\xi_*(\theta^t,E^t), E^t\big) + n^t \quad \quad n^t \sim \mathcal{N}(0,N^t)
\end{equation}
where $\xi_*$ is the true correction the human intends to provide, and $n^t$ is the \emph{observation noise} at iteration $t$. We again assume that $n^t$ is drawn from a zero-mean Gaussian distribution with covariance $N^t$. Here observation noise indicates that actual end-users are unable to provide exactly the feature counts that they have in mind: e.g., it is challenging to perfectly guide a multi-DoF manipulator \cite{akgun2012}.

\textbf{Biased Feedback}. Although \eref{eq:K2} assumes that the observation noise is unbiased, this simplification may not always be true in practice. We therefore perform simulations where the human's feedback is biased in Section~\ref{sec:sim} to support our Kalman filter approach for cases where $n^t$ is not Gaussian noise.

\textbf{Intended Correction}. In \eref{eq:K2}, $\xi_*$ is the correction the human intends to provide given that their current preferences are $\theta^t$ and the environment is $E^t$. Many choices of $\xi_*$ are possible. To be consistent with previous works, we here assume that the human intends to give the following correction:
\begin{equation} \label{eq:K3}
	\xi_*(\theta^t,E^t) = \text{arg}\max_{\xi \in \Xi} ~ \theta^t \cdot \phi(\xi,E^t)
\end{equation} 
Recalling \eref{eq:B6}, notice that $\xi_*$ is now the \emph{optimal trajectory} that maximizes $R$. We recognize that modeling the human as intending to provide the optimal trajectory \eref{eq:K3} and then incorporating Gaussian observation noise over the feature counts \eref{eq:K2} \emph{is analogous to noisy optimal demonstrations} \cite{ziebart2008, ramachandran2007}.

\textbf{IRL as a Dynamical System}. Together \eref{eq:K1}--\eref{eq:K3} express iterative IRL as a dynamical system, where the human's preferences are noisily constant, and the human demonstrates approximately optimal feature counts as a function of the their hidden preferences. We want to estimate these preferences.

\textbf{Extended Kalman Filter}. Given the transition model \eref{eq:K1} and observation model \eref{eq:K2}, we can leverage a Kalman filter to obtain an optimal estimate of the human's preferences $\theta^t$ \cite{choset2005}. To be more precise, since the observation model is here nonlinear, we apply an \emph{extended Kalman filter} (EKF). This EKF linearizes the observation model around the current estimated preferences, and then acts as a standard Kalman filter. We point out that recent developments, such as the \emph{unscented Kalman filter} (UKF) \cite{wan2000}, may outperform an EKF \cite{kandepu2008}. For simplicity of exposition---as well as the insight it provides---we here present the EKF, while noting that we implement the UKF in our simulations.

\textbf{Preference Estimate and Covariance}. Let $\hat{\theta}^t$ be the \emph{mean estimate} of the human's preferences, and let $P^t$ be the \emph{covariance} (i.e., uncertainty) \emph{of this estimate}. Here we list the steps to update the estimate and covariance matrix via an EKF. First, we use a Taylor series expansion to linearize the observation model \eref{eq:K2} around the current estimate, and reach the observation Jacobian $H \in \mathbb{R}^{k \times k}$:
\begin{equation} \label{eq:K4}
	H^t = \frac{\partial \phi}{\partial \xi_*}\cdot \frac{\partial \xi_*}{\partial \theta} \bigg|_{\hat{\theta}^t}
\end{equation}
Intuitively, $H$ tells us how the intended feature counts will vary as the human's preferences change. We expect the performance of our EKF to improve when \eref{eq:K4} is approximately linear. Applying $H$, we can now write the EKF update rule for iterative IRL:
\begin{equation} \label{eq:K45}
	\hat{\theta}^{t+1} = \hat{\theta}^t + K^t\big[\phi(\xi_h^t,E^t) - \phi(\xi_*(\hat{\theta}^t,E^t),E^t)\big]
\end{equation}
Because $\xi_*$ within \eref{eq:K45} is the optimal trajectory given $\hat{\theta}^t$ and $E^t$, we see that $\xi_* = \xi^t$ from \eref{eq:B6}. Therefore, we substitute $\xi_* = \xi^t$ into \eref{eq:K45} to finally derive our \emph{proposed update rule}\footnote{Read $\hat{\theta}^{t+1}$ as the mean estimate of $\theta$ at iteration $t$, given the observed feature counts after $t$ iterations.}:
\begin{equation} \label{eq:K5}
	\hat{\theta}^{t+1} = \hat{\theta}^t + K^t\big[\phi(\xi^t_h,E^t) - \phi(\xi^t,E^t)\big]
\end{equation}
Comparing this proposed update rule \eref{eq:K5} to the Preference Perceptron \eref{eq:B5}, we have straightforwardly replaced the \emph{learning rate} $\alpha > 0$ with the \emph{Kalman gain matrix} $K \in \mathbb{R}^{k \times k}$:
\begin{equation} \label{eq:K6}
	K^t = (P^t + M^t)(H^t)^T\big[H^t(P^t + M^t)(H^t)^T + N^t\big]^{-1}
\end{equation}
Since we are now using a Kalman filter, however, we additionally obtain the \emph{covariance matrix of the estimate}, $P \in \mathbb{R}^{k \times k}$, which is updated according to:
\begin{equation} \label{eq:K7}
	P^{t+1} = (I - K^t H^t)(P^t + M^t)
\end{equation}

\textbf{Summary}. After formulating iterative IRL as a dynamical system with state $\theta$, we derived a Kalman filter estimate of the human's preferences. This approach has \emph{extended} the Preference Perceptron \eref{eq:B5} by adding the Kalman gain matrix, $K$ from \eref{eq:K6}, and the covariance matrix, $P$ from \eref{eq:K7}. Our resulting update rule \eref{eq:K5} is proportional to the Preference Perceptron \eref{eq:B5}, but now we have additionally obtained the covariance (i.e., the uncertainty) of the estimated human preferences \eref{eq:K7}.

\section{Leveraging Uncertainty when Learning from Corrections} \label{sec:apply}

We showed that the Preference Perceptron can be extended to include uncertainty via a Kalman filter: but how should we leverage this uncertainty? In this section, we explore how the covariance of the robot's estimate, $P$, can be used to \emph{actively learn} from human corrections, and then \emph{safely deploy} a resultant trajectory. We consider examples consistent with previous applications of learning from corrections \cite{bajcsy2017, bajcsy2018, jain2015} where: (a) the robot can select virtual environments to elicit more informative human corrections, and (b) the robot deploys with risk-averse or risk-sensitive behavior.

\textbf{Minimizing Covariance}. One reasonable goal for a robot that is learning the human's preferences $\theta$ is to \emph{minimize its uncertainty over those preferences}. When uncertainty is high, the robot is unsure about how it should behave, and when uncertainty is low, the robot is confident that it understands the human's preferences. Within our Kalman filter approach, the robot should therefore elicit corrections that minimize the covariance matrix $P$. Eliciting these corrections is an instance of active learning.

\begin{figure}[t]

	\begin{center}
		\includegraphics[width=1.0\columnwidth]{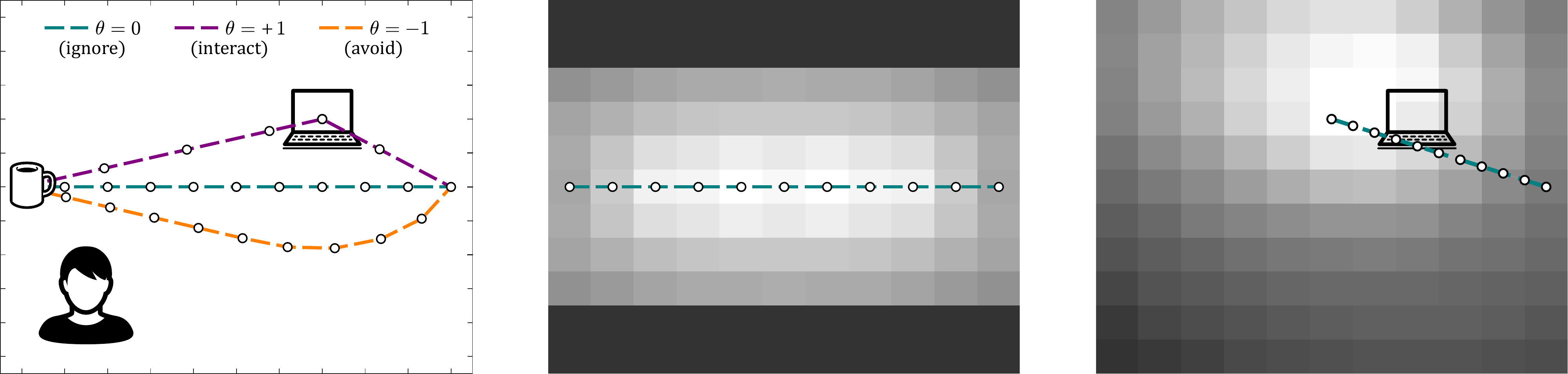}
        
        \vspace{1em}

		\caption{Selecting an environment to minimize uncertainty. \textbf{Left}: The robot is carrying a cup of coffee to a goal location, while a human stands nearby. The robot does not know whether it should carry the coffee over the user's laptop (three possible preferences are given). \textbf{Middle}: The robot's current trajectory $\xi$ is shown. We consider where best to place the laptop to \emph{minimize the robot's covariance} $P$ (prior to the human's correction). Lighter grid cells indicate laptop positions that will minimize $P$. \textbf{Right}: Alternatively, we could fix the laptop location (as shown), and then vary the robot's start location. Lighter grid cells indicate start locations that will best minimize the robot's covariance. We used a UKF \cite{wan2000} and TrajOpt \cite{schulman2014} to perform these simulations.}

		\label{fig:sim1}
	\end{center}

\end{figure}

\textbf{Active Learning}. The robot can elicit corrections that reduce uncertainty by altering aspects of the environment $E \in \mathcal{E}$ in which those corrections are provided. For instance, the robot might change its start state. Alternatively---because the robot is learning from corrections---we can use simulated (i.e., virtual) environments for learning \cite{jain2015}. We here perform simulations for both settings: some where only the start state can be changed, and others where the virtual environment can be altered.

\textbf{Greedy Start and/or Environment Selection}. At each iteration $t$, the robot greedily minimizes its uncertainty \emph{regardless of the human's actual correction} by selecting $E^t \in \mathcal{E}$:
\begin{equation} \label{eq:A1}
	E^t = \text{arg}\min_{E \in \mathcal{E}} ~\| P^{t+1}\|_F = \text{arg}\min_{E \in \mathcal{E}} ~\| (I - K^t H^t)(P^t + M^t) \|_F
\end{equation}
In the above, $\| \cdot \|_F$ is the Frobenius norm (although other norms can be used). Note that $H$ depends on $E$ from \eref{eq:K4}, and so $K$ also depends on $E$ from \eref{eq:K6}. As pointed out by \cite{berg2011}, we can evaluate the uncertainty $P^{t+1}$ in advance---i.e., before the human provides a correction---and hence we can solve \eref{eq:A1} to select an environment \emph{without knowing what correction the human will actually give}.

\textbf{Intuition}. Fig.~\ref{fig:sim1} demonstrates how we can leverage greedy environment selection. Inspecting these results, we see that environments where small changes in $\theta$ lead to large changes in $\phi$ better reduce uncertainty; put another way, we generally want to \emph{maximize} $H$. For example, consider the middle simulation in Fig.~\ref{fig:sim1}. When the laptop is too far away from the robot's current optimal trajectory $\xi$, local corrections do not alter the feature counts, and so the robot cannot learn from this environment.

\begin{figure}[t]

	\begin{center}
		\includegraphics[width=1.0\columnwidth]{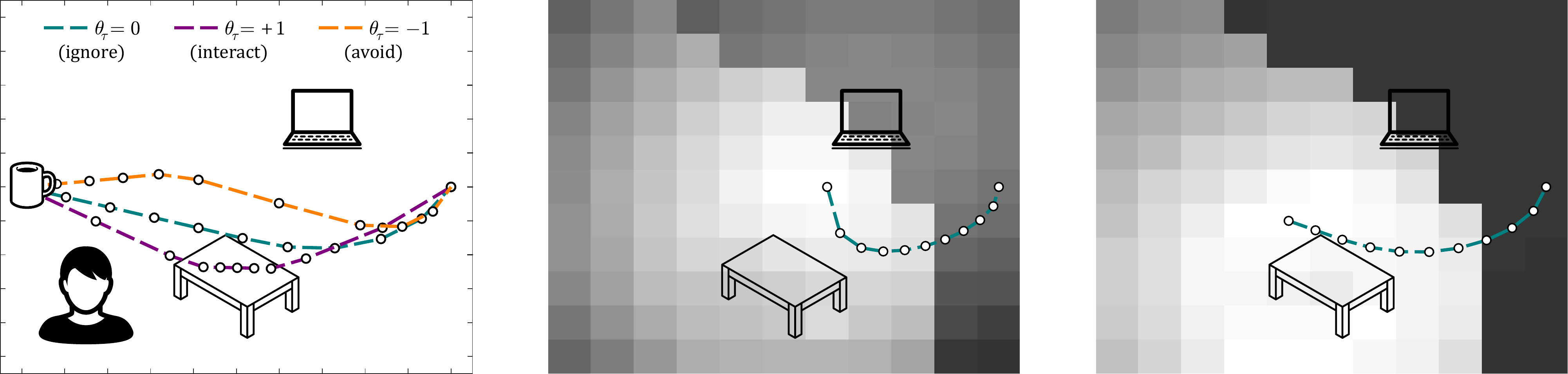}
        
        \vspace{1em}

		\caption{Selecting an environment to minimize uncertainty with multiple features. \textbf{Left}: The task is the same as Fig~\ref{fig:sim1}, but now with a table which the human might prefer for the robot to move across. While the robot has previously learned to avoid the laptop, it does not know $\theta_{\tau}$, the human's true preference for the table. \textbf{Middle}: Assuming that the laptop and table have a fixed position, the robot searches for the best start location. Here the initial covariance is the same for both features. \textbf{Right}: Next, we increase the initial covariance over $\hat{\theta}_{\tau}$. This is meant to emulate situations where \emph{one feature is well understood}, while the robot is \emph{uncertain about another feature}. As before, lighter grid cells will minimize the robot's uncertainty, and we used a UKF with TrajOpt.}

		\label{fig:sim2}
	\end{center}

\end{figure}

\textbf{Multiple Features}. We next use \eref{eq:A1} to select informative environments when the robot is uncertain about multiple features; here the robot must trade-off between learning different features (see Fig.~\ref{fig:sim2}). We find that---if the covariance over each feature is equal---interacting with all features is optimal. On the other hand, when the robot has greater uncertainty over a specific feature, the greedy robot favors environments that elicit corrections on that feature. In Fig.~\ref{fig:sim2}, the robot chooses a start location \emph{between} the laptop and table when the initial uncertainty is equal, but biases its starting location \emph{towards the table} when it has greater uncertainty about the table feature. A robot using \eref{eq:A1} will select environments where the current robot trajectory \emph{interacts with the most uncertain features}.

\textbf{Risk-Sensitive Deployment}. After the robot has learned from the human's corrections and is deployed to perform the task (without human feedback), we can leverage the covariance matrix $P^t$ to select \emph{safer} robotic behavior. Recall that the robot's trajectory optimizes \eref{eq:B6} based on the estimated preferences $\hat{\theta}^t$. Planning only with $\hat{\theta}^t$ fails to account for the covariance over this estimate: the robot might be confident about some learned preferences, but unsure about others. Hence, we will use a \emph{risk-averse} trajectory planning approach similar to \cite{hadfield2017}. First, we generate a set of preferences $\Gamma^t$:
\begin{equation} \label{eq:A2}
	\Gamma_0^t = \hat{\theta}^t\,; \quad \Gamma_i = \hat{\theta}^t + \big(\sqrt{P^t}\big)_i, \enskip i = 1,\ldots,k; \quad \Gamma_i = \hat{\theta}^t - \big(\sqrt{P^t}\big)_{i-k}, \enskip i = k+1,\ldots,2k
\end{equation}
Here $\big(\sqrt{P^t}\big)_i$ is the $i$-th column of the matrix square root of $P^t$. The robot now has $2k+1$ estimates of $\theta$, where $\Gamma_0^t$ is the Kalman filter estimate, and $\Gamma_{1:2k}^t$ are one standard deviation away (as defined by the current covariance $P^t$). Our risk-averse robot optimizes the worst-case reward over $\Gamma^t$:
\begin{equation} \label{eq:A3}
	\xi^t = \text{arg}\max_{\xi \in \Xi} \bigg\{\min_{\gamma \in \Gamma^t}~ \gamma \cdot \phi(\xi, E^t) \bigg\}
\end{equation}
By extension, a \emph{risk-seeking} robot optimizes the best-case reward over $\Gamma^t$, i.e., uses $\max$ in \eref{eq:A3}, and a \emph{risk-neutral} robot simply optimizes with the mean estimate $\Gamma^t_0$, which reduces to \eref{eq:B6}.

\begin{figure}[t]

	\begin{center}
		\includegraphics[width=1.0\columnwidth]{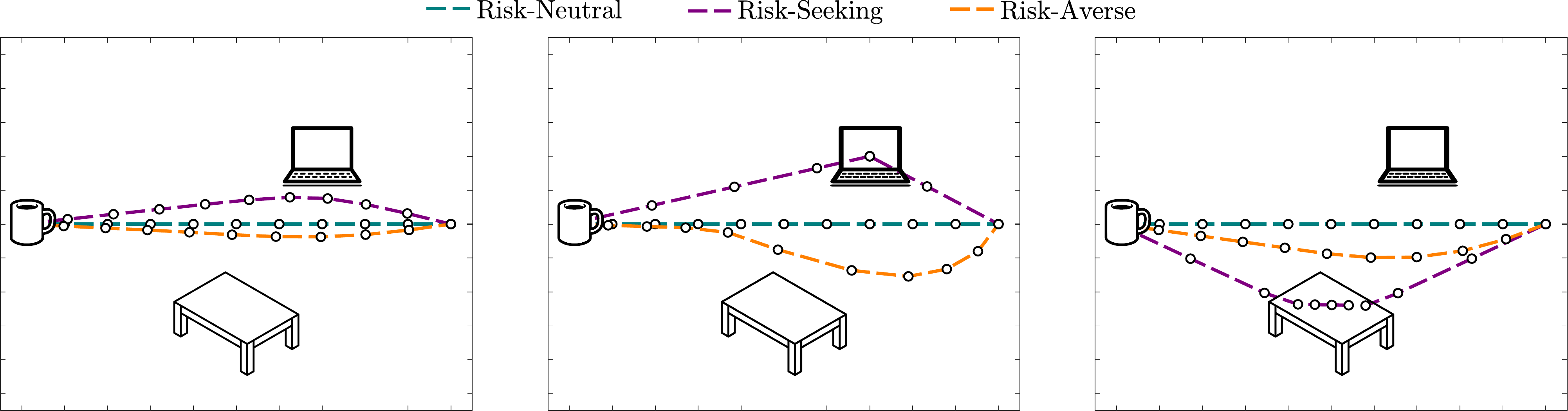}
        
        \vspace{1em}

		\caption{Risk-sensitive deployment based on uncertainty. The robot is performing the same task as in Fig.~\ref{fig:sim2}, but now without human supervision. \textbf{Left}: If the robot has little uncertainty, risk-sensitive planning is almost the same as risk-neutral planning. \textbf{Middle}: When we increase the uncertainty, the robot attempts to \emph{decrease} feature counts (risk-averse) or \emph{increase} feature counts (risk-seeking) as compared to the learned preferences (risk-neutral). \textbf{Right}: Previously, the uncertainty over both the table and laptop features was equivalent. Here the robot is confident about the laptop feature, but uncertain as to whether it should avoid crossing the table. We used \eref{eq:A2} and \eref{eq:A4} together with TrajOpt to perform these simulations in a continuous state space.}

		\label{fig:sim3}
	\end{center}

\end{figure}

\textbf{Simplification}. In practice, solving \eref{eq:A3} for multi-DoF robotic manipulators moving in continuous spaces is challenging. One particular concern is local minima, which naturally occur during trajectory optimization \cite{schulman2014, pan2014}; this problem is now compounded in \eref{eq:A3} by a nested optimization. To make risk-sensitive planning more tractable, we will \emph{reverse} the order of optimization:
\begin{equation} \label{eq:A4}
	\gamma^t = \text{arg}\min_{\gamma \in \Gamma^t}\bigg\{\max_{\xi \in \Xi}~ \gamma \cdot \phi(\xi, E^t) \bigg\}
\end{equation}
In the above, we first find the best possible reward for each preference in $\Gamma^t$, and then choose the worst-case preference $\gamma^t$. Finally, we use $\gamma^t = \hat{\theta}^t$ in \eref{eq:B6}, and obtain the risk-adverse trajectory. We demonstrate the results of risk-sensitive deployment using this simplification in Fig.~\ref{fig:sim3}.

\textbf{Summary}. We leveraged the robot's uncertainty during both \emph{learning} and \emph{deployment}. We demonstrated how the robot could actively learn by adjusting its start state (or, more generally, the environment) to elicit corrections from the human that reduce the robot's uncertainty \eref{eq:A1}. We intuitively found that this active learning caused the robot to interact with the features about which it was most unsure (see Figs.~\ref{fig:sim1} and \ref{fig:sim2}). Next, when the robot is deployed---i.e., no more human corrections are provided---we showed how the robot could exploit uncertainty during risk-sensitive planning. We described a simplified approach for risk-sensitive planning in \eref{eq:A2} and \eref{eq:A4} that is tractable for continuous state spaces. Risk-averse planning using this approach resulted in robots that avoided interacting with preferences that were not clearly understood (see Fig.~\ref{fig:sim3}, risk-averse case). 

Thus, when the robot knows what it does not know, \emph{the robot explores preferences with high uncertainty while learning from human corrections} (learning), \emph{and then avoids preferences with high uncertainty after the human stops providing corrections} (deployment).

\section{Learning Simulations} \label{sec:sim}

In order to support our proposed Kalman filter approach (Section~\ref{sec:kalman}) and demonstrate its active learning application (Section~\ref{sec:apply}), we here perform user simulations where the robot iteratively learns from human corrections. We compare robots that learn with the \emph{Preference Perceptron (PP)}, robots that learn with our \emph{Kalman Filter (KF)} approach, and robots that leverage this Kalman filter for \emph{Active Learning (AL)}. The simulated human end-user provides a correction at each iteration: importantly, these imperfect corrections are biased, and violate our Gaussian observation noise assumption from \eref{eq:K2}. We hypothesize that---even though the simulated human behavior does not match our Kalman filter assumptions---robots that obtain and reason over uncertainty (KF and AL) will learn from human corrections faster than the state-of-the-art (PP).

\textbf{Setup}. The robotic manipulator is carrying a cup of coffee, and is unsure whether the user would prefer for the robot to move over a laptop and/or across a table (see Fig.~\ref{fig:sim2}). At each iteration $t$, the robot observes an environment $E^t$ and executes the optimal trajectory given that environment and $\hat{\theta}^t$, its current estimate of the user's preferences. There are $48$ possible environments $E \in \mathcal{E}$: these environments have different start states, laptop locations, and table locations. The KF and PP robots are given environments \emph{uniformly at random}, while the AL robot \emph{greedily selects} $E^t$ using \eref{eq:A1}.

\textbf{Biased Users}. The simulated human end-user observes $E^t$ and corrects the robot's trajectory $\xi^t$ at each iteration. This realistic user does not provide optimal or noisily optimal corrections; instead, the human's corrections are biased improvements. More specifically, the simulated human corrects the robot's trajectory, $\xi^t$, by moving \emph{one waypoint} from $\xi^t$ towards the equivalent waypoint along their intended trajectory, $\xi_*^t$. The human corrects only the waypoint with the largest error. We emphasize that the simulated human therefore violates our Gaussian noise assumption from \eref{eq:K2}, and tests the robustness of our Kalman filter approach within a realistic learning from corrections setting \cite{jain2015}.

\textbf{Implementation}. The PP and KF robots were simulated $100$ times to obtain their expected performance across randomly selected environments; since the AL robot deterministically selects environments, it was only simulated once. To ensure that the learning rate $\alpha$ for PP was consistent with the Kalman gain $K$ for KF and AL, we set $\alpha^t$ as the expected mean value of the matrix diagonal of $K^t$ across all KF simulations. Like before, we used TrajOpt \cite{schulman2014} to obtain the optimal robot trajectory $\xi^t$, and we used an unscented Kalman filter (UFK) \cite{wan2000} for the KF and AL robots.

\begin{figure}[t]

	\begin{center}
		\includegraphics[width=0.825\columnwidth]{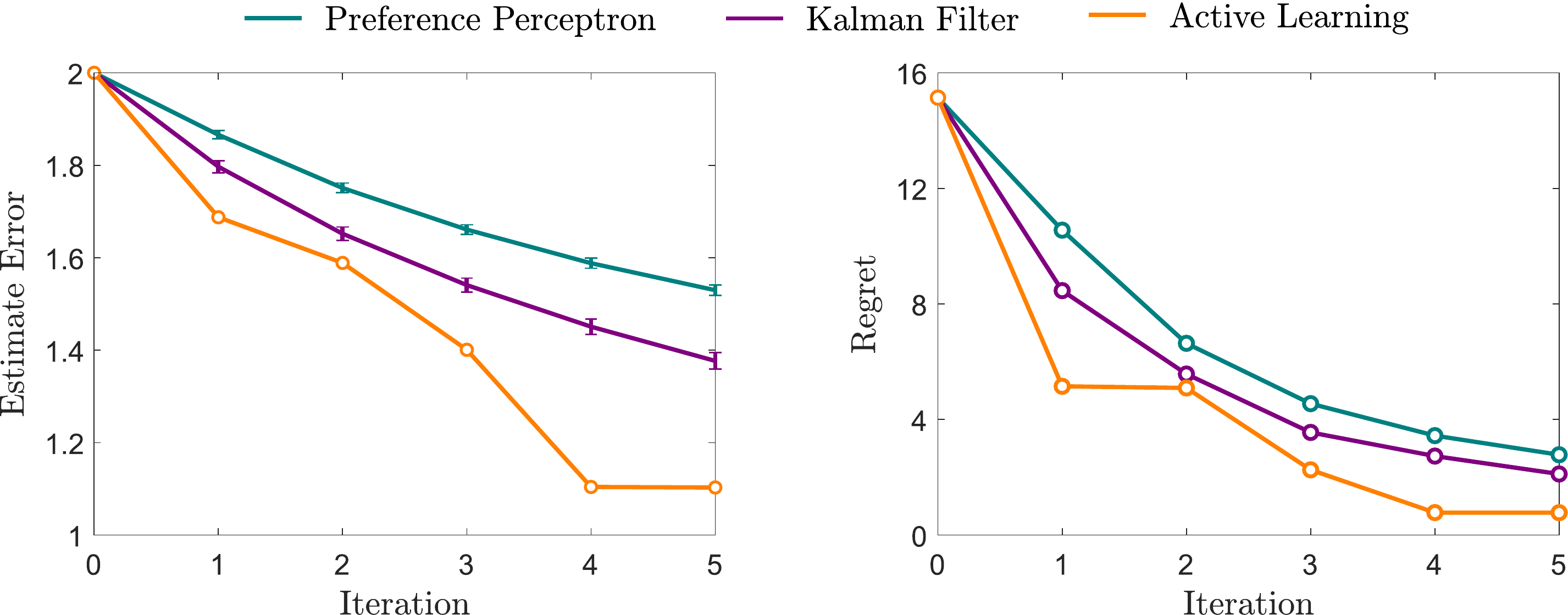}
        
        \vspace{1em}

		\caption{Comparing PP, KF, and AL when iteratively learning from corrections. In PP, the robot learns a maxmium a posteriori estimate. In KF, the robot also obtains the uncertainty of this estimate. In AL, we use the uncertainty to elicit more informative corrections. Here the true preferences are $\theta = (+1,-1)$, and the robot's initial estimate is $\hat{\theta}^0 = (0,0)$. The initial covariance is equal over both features, such that $P^0 = I$. Error bars show standard error of the mean. We found \emph{Regret} using the expected values of $\hat{\theta}^t$ for PP and KF.}

		\label{fig:sim4}
	\end{center}

\end{figure}

\begin{figure}[t]

	\begin{center}
		\includegraphics[width=0.825\columnwidth]{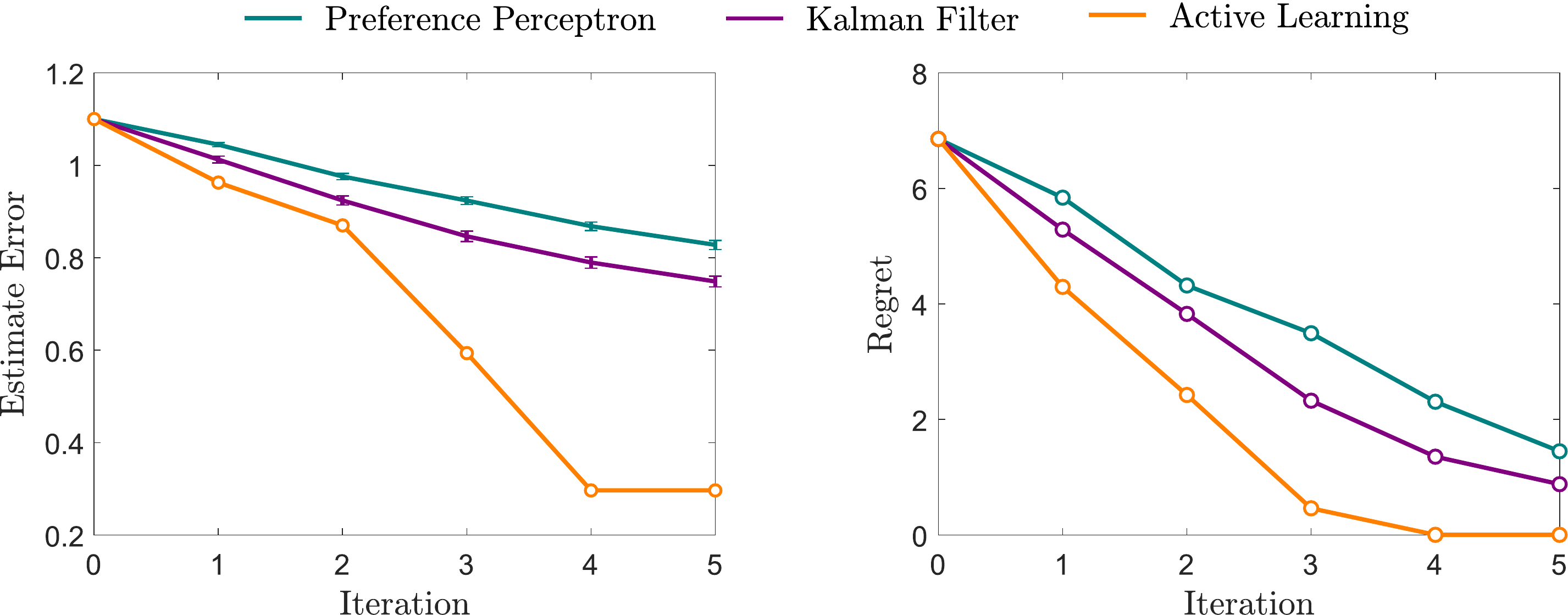}
        
        \vspace{1em}

		\caption{Comparing PP, KF, and AL when iteratively learning from human corrections. Unlike Fig.~\ref{fig:sim4}, here the robot starts with an accurate estimate of $\theta_c$, the human's true preference for the coffee, but a poor estimate of $\theta_{\tau}$, the human's true preference for the table. The true preferences are still $(\theta_c,\theta_{\tau}) = (+1,-1)$, but the robot's initial estimate is $\hat{\theta}^0 = (+0.9,0)$. The robot's initial covariance over the table weight is higher than the initial covariance over the coffee weight, such that $P^0 = \text{diag}(10^{-2},1)$.}

		\label{fig:sim5}
	\end{center}

\end{figure}

\textbf{Results}. Our results are summarized in Figs.~\ref{fig:sim4} and \ref{fig:sim5}. Within these plots, \emph{Estimate Error} refers to the difference between the true preferences, $\theta$, and the robot's estimate, $\hat{\theta}$. We calculated this metric with the $L_1$ norm: ${\|\hat{\theta}^t - \theta\|_1}$. \emph{Regret} captures the difference between the reward the robot would receive if it knew $\theta$, and the reward the robot actually receives using its estimate $\hat{\theta}$. \emph{Regret} is found by comparing $\xi_*^t$, the optimal trajectory given $\theta$, and $\xi_t$, the optimal trajectory given $\hat{\theta}^t$. Recalling \eref{eq:B1}, \emph{Regret} equals: $\theta \cdot \phi(\xi_*^t,E^t) - \theta \cdot \phi(\xi^t,E^t)$. We summed \emph{Regret} across all environments $E \in \mathcal{E}$.

\textbf{Discussion}. Based on our results, KF and AL outperformed the state-of-the-art PP in terms of both \emph{Estimate Error} and \emph{Regret}. From Fig.~\ref{fig:sim4}, we found that KF and AL led to faster learning than PP when the robot's initial uncertainty over the human's preferences was uniform. In Fig.~\ref{fig:sim5}, we found that AL was especially advantageous when some aspects of the human's preferences were initially well understood, but the robot was uncertain about others. These results also demonstrate that our Kalman filter approach is effective even when the human corrections are biased.

\section{Conclusion}

When learning from human corrections, we argue that the robot should recognize which preferences are well understood, and which user preferences remain uncertain. We therefore proposed a Kalman filter approach to iterative IRL, which extends the state-of-the-art by obtaining the uncertainty over the learned human preferences. We then demonstrated how the robot can leverage (a) active learning to reduce this uncertainty when learning from human corrections, and (b) risk-sensitive deployment to avoid uncertain preferences after the human stops providing corrections. Our user simulations showed that the proposed approach results in faster learning than the current state-of-the-art, even for cases where the biased human corrections do not match our Kalman filter assumptions.






\clearpage
\acknowledgments{We would like to thank the reviewers for their thoughtful and encouraging advice. We would also like to thank Scott Niekum for reaching out and bringing related work to our attention. 

This project was funded in part by the NSF GRFP-1450681. The authors are both members of the Mechatronics and Haptic Interfaces (MAHI) Laboratory, Department of Mechanical Engineering, Rice University, Houston, TX 77005.}


\bibliography{citations}  

\begin{thebibliography}{26}
\providecommand{\natexlab}[1]{#1}
\providecommand{\url}[1]{\texttt{#1}}
\expandafter\ifx\csname urlstyle\endcsname\relax
  \providecommand{\doi}[1]{doi: #1}\else
  \providecommand{\doi}{doi: \begingroup \urlstyle{rm}\Url}\fi

\bibitem[Ratliff et~al.(2006)Ratliff, Bagnell, and Zinkevich]{ratliff2006}
N.~D. Ratliff, J.~A. Bagnell, and M.~A. Zinkevich.
\newblock Maximum margin planning.
\newblock In \emph{Proc. International Conference on Machine Learning (ICML)},
  pages 729--736, 2006.

\bibitem[Shivaswamy and Joachims(2015)]{shivaswamy2015}
P.~Shivaswamy and T.~Joachims.
\newblock Coactive learning.
\newblock \emph{Journal of Artificial Intelligence Research}, 53:\penalty0
  1--40, 2015.

\bibitem[Abbeel and Ng(2004)]{abbeel2004}
P.~Abbeel and A.~Y. Ng.
\newblock Apprenticeship learning via inverse reinforcement learning.
\newblock In \emph{Proc. International Conference on Machine Learning (ICML)},
  2004.

\bibitem[Ng and Russell(2000)]{ng2000}
A.~Y. Ng and S.~J. Russell.
\newblock Algorithms for inverse reinforcement learning.
\newblock In \emph{Proc. International Conference on Machine Learning (ICML)},
  pages 663--670, 2000.

\bibitem[Osa et~al.(2018)Osa, Pajarinen, Neumann, Bagnell, Abbeel, and
  Peters]{osa2018}
T.~Osa, J.~Pajarinen, G.~Neumann, J.~A. Bagnell, P.~Abbeel, and J.~Peters.
\newblock An algorithmic perspective on imitation learning.
\newblock \emph{Foundations and Trends in Robotics}, 7\penalty0 (1-2):\penalty0
  1--179, 2018.

\bibitem[Akgun et~al.(2012)Akgun, Cakmak, Jiang, and Thomaz]{akgun2012}
B.~Akgun, M.~Cakmak, K.~Jiang, and A.~L. Thomaz.
\newblock Keyframe-based learning from demonstration.
\newblock \emph{International Journal of Social Robotics}, 4\penalty0
  (4):\penalty0 343--355, 2012.

\bibitem[Ziebart et~al.(2008)Ziebart, Maas, Bagnell, and Dey]{ziebart2008}
B.~D. Ziebart, A.~L. Maas, J.~A. Bagnell, and A.~K. Dey.
\newblock Maximum entropy inverse reinforcement learning.
\newblock In \emph{Proc. Association for the Advancement of Artificial
  Intelligence (AAAI)}, volume~8, pages 1433--1438, 2008.

\bibitem[Ramachandran and Amir(2007)]{ramachandran2007}
D.~Ramachandran and E.~Amir.
\newblock Bayesian inverse reinforcement learning.
\newblock \emph{Urbana}, 51\penalty0 (61801):\penalty0 1--4, 2007.

\bibitem[Bajcsy et~al.(2017)Bajcsy, Losey, O'Malley, and Dragan]{bajcsy2017}
A.~Bajcsy, D.~P. Losey, M.~K. O'Malley, and A.~D. Dragan.
\newblock Learning robot objectives from physical human interaction.
\newblock In \emph{Prof. Conference on Robot Learning (CoRL)}, pages 217--226,
  2017.

\bibitem[Bajcsy et~al.(2018)Bajcsy, Losey, O'Malley, and Dragan]{bajcsy2018}
A.~Bajcsy, D.~P. Losey, M.~K. O'Malley, and A.~D. Dragan.
\newblock Learning from physical human corrections, one feature at a time.
\newblock In \emph{Proc. ACM/IEEE International Conference on Human-Robot
  Interaction (HRI)}, pages 141--149, 2018.

\bibitem[Jain et~al.(2015)Jain, Sharma, Joachims, and Saxena]{jain2015}
A.~Jain, S.~Sharma, T.~Joachims, and A.~Saxena.
\newblock Learning preferences for manipulation tasks from online coactive
  feedback.
\newblock \emph{The International Journal of Robotics Research}, 34\penalty0
  (10):\penalty0 1296--1313, 2015.

\bibitem[Settles(2012)]{settles2012}
B.~Settles.
\newblock Active learning.
\newblock \emph{Synthesis Lectures on Artificial Intelligence and Machine
  Learning}, 6\penalty0 (1):\penalty0 1--114, 2012.

\bibitem[Lopes et~al.(2009)Lopes, Melo, and Montesano]{lopes2009}
M.~Lopes, F.~Melo, and L.~Montesano.
\newblock Active learning for reward estimation in inverse reinforcement
  learning.
\newblock In \emph{Proc. Joint European Conference on Machine Learning and
  Knowledge Discovery in Databases (ECML-PKDD)}, pages 31--46, 2009.

\bibitem[Cohn et~al.(2011)Cohn, Durfee, and Singh]{cohn2011}
R.~Cohn, E.~Durfee, and S.~Singh.
\newblock Comparing action-query strategies in semi-autonomous agents.
\newblock In \emph{Proc. International Conference on Autonomous Agents and
  Multiagent Systems (AAMAS)}, pages 1287--1288, 2011.

\bibitem[Cui and Niekum(2018)]{cui2018}
Y.~Cui and S.~Niekum.
\newblock Active reward learning from critiques.
\newblock In \emph{Proc. IEEE International Conference on Robotics and
  Automation (ICRA)}, 2018.

\bibitem[Huang et~al.(2017)Huang, Held, Abbeel, and Dragan]{huang2017}
S.~H. Huang, D.~Held, P.~Abbeel, and A.~D. Dragan.
\newblock Enabling robots to communicate their objectives.
\newblock In \emph{Proc. Robotics: Science and Systems (RSS)}, 2017.

\bibitem[Puterman(2014)]{puterman2014}
M.~L. Puterman.
\newblock \emph{Markov Decision Processes: Discrete Stochastic Dynamic
  Programming}.
\newblock John Wiley \& Sons, 2014.

\bibitem[Bottou(1998)]{bottou1998}
L.~Bottou.
\newblock Online learning and stochastic approximations.
\newblock In \emph{On-line Learning in Neural Networks}, volume~17, pages
  9--42, 1998.

\bibitem[Karaman and Frazzoli(2011)]{karaman2011}
S.~Karaman and E.~Frazzoli.
\newblock Sampling-based algorithms for optimal motion planning.
\newblock \emph{The International Journal of Robotics Research}, 30\penalty0
  (7):\penalty0 846--894, 2011.

\bibitem[Schulman et~al.(2014)Schulman, Duan, Ho, Lee, Awwal, Bradlow, Pan,
  Patil, Goldberg, and Abbeel]{schulman2014}
J.~Schulman, Y.~Duan, J.~Ho, A.~Lee, I.~Awwal, H.~Bradlow, J.~Pan, S.~Patil,
  K.~Goldberg, and P.~Abbeel.
\newblock Motion planning with sequential convex optimization and convex
  collision checking.
\newblock \emph{The International Journal of Robotics Research}, 33\penalty0
  (9):\penalty0 1251--1270, 2014.

\bibitem[Choset(2005)]{choset2005}
H.~M. Choset.
\newblock \emph{Principles of Robot Motion: Theory, Algorithms, and
  Implementation}.
\newblock MIT press, 2005.

\bibitem[Wan and Van Der~Merwe(2000)]{wan2000}
E.~A. Wan and R.~Van Der~Merwe.
\newblock The unscented {K}alman filter for nonlinear estimation.
\newblock In \emph{Proc. Adaptive Systems for Signal Processing,
  Communications, and Control Symposium (AS-SPCC)}, pages 153--158, 2000.

\bibitem[Kandepu et~al.(2008)Kandepu, Foss, and Imsland]{kandepu2008}
R.~Kandepu, B.~Foss, and L.~Imsland.
\newblock Applying the unscented kalman filter for nonlinear state estimation.
\newblock \emph{Journal of Process Control}, 18\penalty0 (7-8):\penalty0
  753--768, 2008.

\bibitem[Van Den~Berg et~al.(2011)Van Den~Berg, Abbeel, and Goldberg]{berg2011}
J.~Van Den~Berg, P.~Abbeel, and K.~Goldberg.
\newblock {LQG-MP}: {O}ptimized path planning for robots with motion
  uncertainty and imperfect state information.
\newblock \emph{The International Journal of Robotics Research}, 30\penalty0
  (7):\penalty0 895--913, 2011.

\bibitem[Hadfield-Menell et~al.(2017)Hadfield-Menell, Milli, Abbeel, Russell,
  and Dragan]{hadfield2017}
D.~Hadfield-Menell, S.~Milli, P.~Abbeel, S.~J. Russell, and A.~Dragan.
\newblock Inverse reward design.
\newblock In \emph{Proc. Advances in Neural Information Processing Systems
  (NIPS)}, pages 6768--6777, 2017.

\bibitem[Pan et~al.(2014)Pan, Chen, and Abbeel]{pan2014}
J.~Pan, Z.~Chen, and P.~Abbeel.
\newblock Predicting initialization effectiveness for trajectory optimization.
\newblock In \emph{Proc. IEEE International Conference on Robotics and
  Automation (ICRA)}, pages 5183--5190, 2014.

\end{thebibliography}

\end{document}